\documentclass[conference]{IEEEtran}
\IEEEoverridecommandlockouts
\usepackage{cite}
\usepackage{amsmath,amssymb,amsfonts}
\usepackage{algorithm}
\usepackage{algorithmic}
\usepackage{graphicx}
\usepackage{textcomp}
\usepackage{xcolor}
\def\BibTeX{{\rm B\kern-.05em{\sc i\kern-.025em b}\kern-.08em
    T\kern-.1667em\lower.7ex\hbox{E}\kern-.125emX}}
\begin{document}

\title{Batch Monte Carlo Tree Search\\
\thanks{Warm thanks to Rémi Coulom who told me he was simulating sequential PUCT in his implementation of Batch PUCT.

This work was supported in part by the French government under management of Agence Nationale de la Recherche as part of the “Investissements d’avenir” program, reference ANR19-P3IA-0001 (PRAIRIE 3IA Institute).
}
}

\author{\IEEEauthorblockN{Tristan Cazenave}
\IEEEauthorblockA{\textit{LAMSADE} \\
\textit{Universit\'e Paris-Dauphine, PSL, CNRS}\\
\textit{Paris, France}\\
Tristan.Cazenave@dauphine.psl.eu}
}

\maketitle

\begin{abstract}
Making inferences with a deep neural network on a batch of states is much faster with a GPU than making inferences on one state after another. We build on this property to propose Monte Carlo Tree Search algorithms using batched inferences. Instead of using either a search tree or a transposition table we propose to use both in the same algorithm. The transposition table contains the results of the inferences while the search tree contains the statistics of Monte Carlo Tree Search. We also propose to analyze multiple heuristics that improve the search: the $\mu$ FPU, the Virtual Mean, the Last Iteration and the Second Move heuristics. They are evaluated for the game of Go using a MobileNet neural network. 
\end{abstract}


\section{Introduction}

\noindent Monte Carlo Tree Search (MCTS) using a combined policy and value network is the state of the art for complex two-player perfect information games such as the game of Go \cite{silver2018general}. MCTS is also the state of the art for many other games and problems \cite{BrownePWLCRTPSC2012}. We propose multiple optimizations of MCTS in the context of its combination with deep neural networks. With current hardware such as GPU or TPU it is much faster to batch the inferences of a deep neural network rather than to perform them sequentially. We give in this paper MCTS algorithms that make inferences in batches and some heuristics to improve them. We evaluate the search algorithms for the game of Go.

The second section deals with existing work on MCTS for games. The third section presents our algorithms. The fourth section details experimental results.

\section{Monte Carlo Tree Search}

MCTS has its roots in computer Go \cite{Coulom2006}. A theoretically well founded algorithm is UCT \cite{UCT2006}. Dealing with transpositions in UCT was addressed with the UCD algorithm \cite{SaffidineCazenaveMehat2011KBS}. The authors tested various ways to deal with transpositions and gave results for multiple games in the context of General Game Playing (GGP).

The GRAVE algorithm \cite{Cazenave15Grave} is the state of the art in GGP. It uses a transposition table as the core of the tree search algorithm. Entries of the transposition table contain various kind of information such as the statistics on the moves as well as the generalized AMAF statistics. It does not use the UCB bandit anymore but an improvement of RAVE \cite{Gelly2011AI}. 

When combined with neural networks the state of the art MCTS algorithm is PUCT \cite{Silver2016AlphaGo}. It is the current best algorithm for games such as Go \cite{silver2017mastering} and Shogi \cite{silver2018general}. It was used in the AlphaGo program \cite{Silver2016AlphaGo} as well as in its descendants AlphaGo Zero \cite{silver2017mastering} and Alpha Zero \cite{silver2018general}.

The PUCT bandit is:

$$V(s, a)= Q(s,a) + c \times P(s,a) \times \frac{\sqrt{N(s)}}{1+N(s,a)}$$

Where $P(s,a)$ is the probability of move $a$ to be the best moves in state $s$ given by the policy head of the neural network, $N(s)$ is the total number of descents performed in state $s$ and $N(s,a)$ is the number of descents for move $a$ in state $s$.

Many researchers have replicated the Alpha Zero experiments and also use the PUCT algorithm \cite{tian2019elfopengo,pascutto2017leela,gzero,greatfast,cazenave2020polygames}.

\subsection{Parallelization of MCTS}

Three different ways to parallelize UCT were first proposed in 2007 \cite{cazenave2007parallelization}. They were further renamed Root Parallelization, Leaf Parallelization and Tree Parallelization \cite{chaslot2008parallel,cazenave2008parallel}.

Other optimizations of Tree Parallelization such as lock free parallelization were also proposed \cite{Enzenberger2009} as well as scalability studies \cite{segal2010scalability}.

\subsection{Virtual Loss}

A virtual loss enables to make multiple descents of the tree in parallel when the results of the evaluations at the leaves are not yet known and the tree has not yet been updated with these results. It is standard in Tree Parallelization \cite{chaslot2008parallel,cazenave2008parallel}. The principle is very simple since it consists in adding a predefined number of visits to the moves that are played during the tree descent.

Virtual loss is used in most of the Go programs including AlphaGo \cite{Silver2016AlphaGo} and ELF \cite{tian2019elfopengo}.

\subsection{Batched Inferences}

Using batch forwards of the neural network to evaluate leaves of the search tree and find the associated priors given by the policy head is current practice in many game programs \cite{tian2019elfopengo,pascutto2017leela,gzero,greatfast,cazenave2020polygames}. Usually a set of leaves is generated with search then the neural network is run on a single batch of leaves and the results are incorporated into the search tree.

\subsection{First Play Urgency}

Vanilla UCT begins by exploring each arm once before using UCB. This behavior was improved with the First Play Urgency (FPU) \cite{wang2007modifications}. A large FPU value ensures the exploration of each move once before further exploitation of any previously visited move. A small FPU ensures earlier exploitation if the first simulations lead to an urgency larger than the FPU.

In more recent MCTS programs using playouts, FPU was replaced by RAVE \cite{Gelly2011AI} which uses the All Moves As First heuristic so as to order moves before switching gradually to UCT. RAVE was later improved with GRAVE \cite{Cazenave15Grave} which has good results in GGP \cite{brownepractical,sironi2019monte}.

In AlphaGo \cite{Silver2016AlphaGo}, the FPU was revived and is set to zero when the evaluations are between -1 and 1. We name this kind of FPU the constant FPU. It has deficiencies. When the FPU is too high, all possible moves are tried at a node before going further below in the state space and this slows down the search and makes it shallow. When the FPU is too low, the moves after the best move are only tried after many simulations and the search does not explore enough. When the constant is in the middle of the range of values as in AlphaGo, both deficiencies can occur, either when the average of the evaluations is below the constant or is greater than the constant.

In other programs such as ELF \cite{tian2019elfopengo} the FPU is set to the best mean of the already explored moves. This is better than using a constant FPU since the best mean is related to the mean of the other moves. It encourages exploration.

Yet another way to deal with the FPU is to set it to the average mean of the node (using the statistics of all the explored moves). We name this kind of FPU the $\mu$ FPU.

\section{The Batch PUCT Algorithm}

In this section we describe our algorithm and its refinements. We first explain how we deal with trees and transposition table. We then detail our heuristics: the Virtual Mean, the Last Iteration and the Second Move.

\subsection{Trees and Transposition Table}

The principle of Batch MCTS is to simulate a sequential MCTS using batched evaluations. In order to do so it separates the states that have been evaluated from the search tree. The transposition table contains the states that have been evaluated. The search tree is developed as in usual sequential MCTS and when it reaches a leaf it looks up the state in the transposition table. If it is present then it sends back the corresponding evaluation. If it is not in the transposition table it sends back the Unknown value and adds the state to the next batch of states.

At the beginning of Monte Carlo Tree Search (MCTS) were random playouts and poor initial selection of moves to try. It changed a lot with Deep Reinforcement Learning (DRL) which brought strong policies and evaluations. The PUCT algorithm strongly biases the moves to try with the learned policy and performs accurate evaluations at the leaves of the MCTS instead of random playouts.

In early MCTS program it was found very useful to use a transposition table in the search tree in order to reuse the information on the moves to try from previous tree descents that took another path in the tree such as in UCD \cite{SaffidineCazenaveMehat2011KBS}. However reusing this information from different descents and starting a playout at a deeper leaf biases the statistics of the upper part of the tree and does not preserve equity between root moves. It was not a problem in playout based MCTS since the tree policy was much stronger than the playout policy. On the contrary biasing the moves in the context of PUCT can be misleading for the evaluation of the root moves.

MCTS algorithms were designed when finding the interesting moves was difficult and when the descent of the tree cost approximately as much as making a playout. With deep neural networks trained using Deep Reinforcement Learning the evaluation at a leaf is now more costly than the descent of the tree. Moreover the policy provides accurate probabilities for the moves to try. In this context it is interesting to reuse the neural network evaluation as much as possible and we do not have anymore to make many playouts to have a good policy in the tree. The algorithm we propose does reuse more than the standard MCTS the evaluations already made and it is less detrimental than in playout based MCTS since we already have a good policy provided by the neural network.

Bath MCTS increases the number of tree descents for a given budget of inferences compared to usual PUCD with a transposition table and it makes the average evaluations of the moves not biased by the developments from different paths of already developed states.

In our implementation of Batch PUCT we use one transposition table and two trees. The transposition table records for each state that has been given to the neural network the evaluation of the state and the priors for the moves of the state. The first tree records the statistics required to calculate the PUCT bandit and the address of the children that have already been explored. The second tree is a copy of the first tree used to build the next batch of states that will be then given to the neural network.


Algorithm \ref{BatchPUCT} give the main PUCT search algorithm using a transposition table and the two trees. The GetBatch boolean is used to make the distinction between the first tree and the second tree. The first tree is the main search tree while the second tree is only used to build the batches.

Algorithm \ref{UpdateStatistics} gives the usual way of updating the statistics used for the main tree. Algorithm \ref{UpdateStatisticsGet} gives the update of the statistics for the second tree.

The main algorithm is the GetMove algorithm (algorithm \ref{GetMove}). It calls the GetBatch algorithm (algorithm \ref{GetBatch}) that descends the second tree many times in order to fill the batch. It then makes inferences on the built batch and calls the PutBatch algorithm (algorithm \ref{PutBatch}) that put the results of the inferences in the transposition table and then upadtes the main tree. GetBatch, forward and PutBatch are called $B$ times. In the end the GetMove algorithm returns the most simulated move of the main tree.

\subsection{The Virtual Mean}

The standard approach in parallel MCTS is to use a virtual loss to avoid exploring again and again the same moves when no new evaluation is available. We propose the Virtual Mean as an alternative to the virtual loss. The Virtual Mean increases the number of simulation of the move as in the virtual loss but it also adds the mean of the move to the sum of its evaluations in order to have more realistic statistics for the next descent. 

\begin{algorithm}
\begin{algorithmic}[1]
\setlength{\lineskip}{3pt}
\STATE{Function BatchPUCT ($s,GetBatch$)}
\begin{ALC@g}
\IF{isTerminal ($s$)}
\RETURN Evaluation ($s$)
\ENDIF
\IF{$GetBatch$}
\STATE{$t \gets treeBatch$}
\ELSE
\STATE{$t \gets tree$}
\ENDIF
\IF{$s \notin t$}
\IF{$s \notin$ transposition table}
\IF{$GetBatch$}
\STATE{add $s$ to the batch}
\ENDIF
\RETURN Unknown
\ELSE
\STATE{add $s$ to $t$}
\RETURN value ($s$)
\ENDIF
\ENDIF
\STATE{$bestScore = -\infty$}
\FOR{$m \in$ legal moves of $s$}
\STATE{$\mu = FPU$}
\IF{$t.p(s,m) > 0$}
\STATE{$\mu = \frac{t.sum(s,m)}{t.p(s,m)}$}
\ENDIF
\STATE{$bandit = \mu + c \times t.prior(s,m) \times \frac{\sqrt{t.p(s)}}{1+t.p(s,m)}$}
\IF{$bandit > bestScore$}
\STATE{$bestScore \gets bandit$}
\STATE{$bestMove \gets m$}
\ENDIF
\ENDFOR
\STATE{$s_1$ = play ($s, bestMove$)}
\STATE{$res$ = BatchPUCT ($s_1$, $GetBatch$)}
\IF{$GetBatch$}
\STATE{UpdateStatisticsGet ($res,bestMove,s, t$)}
\ELSE
\STATE{UpdateStatistics ($res,bestMove,s,t$)}
\ENDIF
\RETURN $res$
\end{ALC@g}
\end{algorithmic}
\caption{\label{BatchPUCT}The BatchPUCT algorithm}
\end{algorithm}

\begin{algorithm}
\begin{algorithmic}[1]
\setlength{\lineskip}{3pt}
\STATE{Function UpdateStatistics ($res, m, s, t$)}
\begin{ALC@g}
\IF{$res \neq Unknown$}
\STATE{$t.p(s,m) = t.p(s,m) + 1$}
\STATE{$t.sum(s,m) = t.sum(s,m) + res$}
\STATE{$t.p(s) = t.p(s) + 1$}
\STATE{$t.sum(s) = t.sum(s) + res$}
\ENDIF
\end{ALC@g}
\end{algorithmic}
\caption{\label{UpdateStatistics}The UpdateStatistics algorithm for the main tree}
\end{algorithm}

Algorithm \ref{UpdateStatisticsGet} gives the different ways of updating the statistics of a node. The $vl$ variable is the number of virtual losses that are added to a move when it leads to an unknown leaf. A value greater than one will encourage more exploration and will avoid resampling again and again the same unknown leaf. The value is related to the maximum number of samples allowed in the GetBatch algorithm (the variable $N$ in algorithm \ref{GetBatch}). A low value of $N$ will miss evaluations and will not completely fill the batch. A large value of $N$ will better fill the batch but will take more time to do it. Increasing $vl$ enables to fill the batch with more states for the same value of $N$. However a too large value of $vl$ can overlook some states and decrease the number of visited nodes in the main search tree. The $res$ value is the evaluation returned by the tree descent, $m$ is the move that has been tried in the descent, $s$ is the state and $t$ is the second tree.

\begin{algorithm}
\begin{algorithmic}[1]
\setlength{\lineskip}{3pt}
\STATE{Function UpdateStatisticsGet ($res, m, s, t$)}
\begin{ALC@g}
\IF{$res = Unknown$}
\STATE{$\mu = \frac{t.sum(s,m)}{t.p(s,m)}$}
\IF{VirtualLoss}
\STATE{$t.p(s,m) = t.p(s,m) + vl$}
\STATE{$t.p(s) = t.p(s) + vl$}
\ELSIF{VirtualMean}
\STATE{$t.p(s,m) = t.p(s,m) + vl$}
\STATE{$t.sum(s,m) = t.sum(s,m) + vl \times \mu$}
\STATE{$t.p(s) = t.p(s) + vl$}
\STATE{$t.sum(s) = t.sum(s) + vl \times \mu$}
\ENDIF
\ELSE
\STATE{$t.p(s,m) = t.p(s,m) + 1$}
\STATE{$t.sum(s,m) = t.sum(s,m) + res$}
\STATE{$t.p(s) = t.p(s) + 1$}
\STATE{$t.sum(s) = t.sum(s) + res$}
\ENDIF
\end{ALC@g}
\end{algorithmic}
\caption{\label{UpdateStatisticsGet}The UpdateStatisticsGet algorithm}
\end{algorithm}

\begin{algorithm}
\begin{algorithmic}[1]
\setlength{\lineskip}{3pt}
\STATE{Function GetBatch ($s$)}
\begin{ALC@g}
\STATE{$treeBatch \gets tree$}
\STATE{$i \gets 0$}
\WHILE{batch is not filled \AND $i < N$}
\STATE{BatchPUCT ($s,True$)}
\STATE{$i \gets i + 1$}
\ENDWHILE
\end{ALC@g}
\end{algorithmic}
\caption{\label{GetBatch}The GetBatch algorithm}
\end{algorithm}

Algorithm \ref{GetBatch} give the main algorithm to build the batch. In order to present the algorithm simply we assume a copy of the main tree to treeBatch which is then used and modified in order to build the batch. A more elaborate implementation is to separate inside a node the statistics of the main tree and the statistics made during the building of the batch. A global stamp can be used to perform a lazy reinitialization of the batch statistics at each new batch build.

\begin{algorithm}
\begin{algorithmic}[1]
\setlength{\lineskip}{3pt}
\STATE{Function PutBatch ($s,out$)}
\begin{ALC@g}
\FOR{$o \in out$}
\STATE{add $o$ to the Transposition Table}
\ENDFOR
\STATE{$res \gets$ BatchPUCT ($s,False$)}
\WHILE{$res \neq Unknown$}
\STATE{$res \gets$ BatchPUCT ($s,False$)}
\ENDWHILE
\end{ALC@g}
\end{algorithmic}
\caption{\label{PutBatch}The PutBatch algorithm}
\end{algorithm}

\begin{algorithm}
\begin{algorithmic}[1]
\setlength{\lineskip}{3pt}
\STATE{Function GetMove ($s,B$)}
\begin{ALC@g}
\FOR{$i \gets 1$ to $B$}
\STATE{GetBatch($s$)}
\STATE{$out \gets$ Forward ($batch$)}
\STATE{PutBatch ($s,out$)}
\ENDFOR
\STATE{$t \gets tree.node (s)$}
\RETURN $argmax_m (t.p(s,m))$
\end{ALC@g}
\end{algorithmic}
\caption{\label{GetMove}The GetMove algorithm}
\end{algorithm}

\subsection{The Last Iteration}

At the end of the GetMove algorithm, many states are evaluated in the transposition table but have not been used in the tree. In order to gain more information it is possible to continue searching for unused state evaluations at the price of small inacurracies.

The principle is to call the BatchPUCT algorithm with GetBatch as True as long as the number of Unknown values sent back does not reach a threshold.

The descents that end with a state which is not in the transposition table do not change the statistics of the moves since they add the mean of the move using the Virtual Mean. The descents that end with an unused state of the transposition table modify the statistics of the moves and improve them as they include statistics on more states.

The Last Iteration algorithm is given in algorithm \ref{GetMoveLastIteration}. The $U$ variable is the number of visited unknown states before the algorithm stops.

\begin{algorithm}
\begin{algorithmic}[1]
\setlength{\lineskip}{3pt}
\STATE{Function GetMoveLastIteration ($s,B$)}
\begin{ALC@g}
\FOR{$i \gets 1$ to $B$}
\STATE{GetBatch($s$)}
\STATE{$out \gets$ Forward ($batch$)}
\STATE{PutBatch ($s,out$)}
\ENDFOR
\STATE{$nbUnknown \gets 0$}
\STATE{$treeBatch \gets tree$}
\WHILE{$nbUnknown < U$}
\STATE{$res \gets$ BatchPUCT ($s,True$)}
\IF{$res = Unknown$}
\STATE{$nbUnknown \gets nbUnknown + 1$}
\ENDIF
\ENDWHILE
\STATE{$t \gets treeBatch.node (s)$}
\RETURN $argmax_m (t.p(s,m))$
\end{ALC@g}
\end{algorithmic}
\caption{\label{GetMoveLastIteration}The GetMoveLastIteration algorithm}
\end{algorithm}

\subsection{The Second Move Heuristic}

Let $n_1$ be the number of playouts of the most simulated move at the root, $n_2$ the number of playouts of the second most simulated move, $b$ the total budget and $rb$ the remaining budget. If $n_1 > n_2 + rb$, it is useless to perform more playouts beginning with the most simulated move since the most simulated move cannot change with the remaining budget. When the most simulated move reaches this threshold it is more useful to completely allocate $rb$ to the second most simulated move and to take as the best move the move with the best mean when all simulations are finished.

The modifications of the search algorithm that implement the Second Move heuristic are given in algorithm \ref{BatchSecond}. Lines 33-39 modify the best move to try at the root when the most simulated move is unreachable. In this case the second most simulated move is preferred.

Algorithm \ref{GetMoveSecondHeuristic} gives the modifications of the GetMove algorithm for using the Second Move heuristic. Lines 8-17 choose between the most simulated move and the second most simulated move according to their means.

\begin{algorithm}
\begin{algorithmic}[1]
\setlength{\lineskip}{3pt}
\STATE{Function BatchSecond ($s,GetBatch,budget,i,root$)}
\begin{ALC@g}
\IF{isTerminal ($s$)}
\RETURN Evaluation ($s$)
\ENDIF
\IF{$GetBatch$}
\STATE{$t \gets treeBatch$}
\ELSE
\STATE{$t \gets tree$}
\ENDIF
\IF{$s \notin t$}
\IF{$s \notin$ transposition table}
\IF{$GetBatch$}
\STATE{add $s$ to the batch}
\ENDIF
\RETURN Unknown
\ELSE
\STATE{add $s$ to $t$}
\RETURN value ($s$)
\ENDIF
\ENDIF
\STATE{$bestScore = -\infty$}
\FOR{$m \in$ legal moves of $s$}
\STATE{$\mu = FPU$}
\IF{$t.p(s,m) > 0$}
\STATE{$\mu = \frac{t.sum(s,m)}{t.p(s,m)}$}
\ENDIF
\STATE{$bandit = \mu + c \times t.prior(s,m) \times \frac{\sqrt{1+t.p(s)}}{1+t.p(s,m)}$}
\IF{$bandit > bestScore$}
\STATE{$bestScore \gets bandit$}
\STATE{$bestMove \gets m$}
\ENDIF
\ENDFOR
\IF{root}
\STATE{$b \gets highestValue_m (t.p(s,m))$}
\STATE{$b_1 \gets secondHighestValue_m (t.p(s,m))$}
\IF{$b \geq b_1 + budget - i$}
\STATE{$bestMove \gets secondBestMove_m (t.p(s,m))$}
\ENDIF
\ENDIF
\STATE{$s_1$ = play ($s, bestMove$)}
\STATE{$res$ = BatchSecond ($s_1,GetBatch,budget,i,False$)}
\IF{$GetBatch$}
\STATE{UpdateStatisticsGet ($res,bestMove,s,t$)}
\ELSE
\STATE{UpdateStatistics ($res,bestMove,s,t$)}
\ENDIF
\RETURN $res$
\end{ALC@g}
\end{algorithmic}
\caption{\label{BatchSecond}The BatchSecond algorithm}
\end{algorithm}

\begin{algorithm}
\begin{algorithmic}[1]
\setlength{\lineskip}{3pt}
\STATE{Function GetMoveSecondHeuristic ($s,B$)}
\begin{ALC@g}
\STATE{$b \gets size (batch)$}
\FOR{$i \gets 0$ to $B$}
\STATE{GetBatchSecond ($s,B \times b,i \times b$)}
\STATE{$out \gets$ Forward ($batch$)}
\STATE{PutBatchSecond ($out,B \times b,i \times b$)}
\ENDFOR
\STATE{$t \gets tree$}
\STATE{$best \gets bestMove_m (t.p(s,m))$}
\STATE{$\mu = \frac{t.sum(s,best)}{t.p(s,best)}$}
\STATE{$secondBest \gets secondBestMove_m (t.p(s,m))$}
\STATE{$\mu_1 = \frac{t.sum(s,secondBest)}{t.p(s,secondBest)}$}
\IF{$\mu_1 > \mu$}
\RETURN $secondBest$
\ELSE
\RETURN $best$
\ENDIF
\end{ALC@g}
\end{algorithmic}
\caption{\label{GetMoveSecondHeuristic}The GetMoveSecondHeuristic algorithm}
\end{algorithm}

\section{Experimental Results}

Experiments are performed using a MobileNet neural network which is an architecture well fitted for the game of Go \cite{Cazenave2021Mobile,Cazenave2021Improving}. The network has 16 blocks, a trunk of 64 and 384 planes in the inverted residual. It has been trained on the Katago dataset containing games played at a superhuman level.

\subsection{FPU}

We test the constant FPU and the best mean FPU against the $\mu$ FPU. Table \ref{tableFPU} gives the average winrate over 400 games of the different FPU for different numbers of playouts. For example, the first cell means that the constant FPU wins 13.00\% of its games against the $\mu$ FPU when the search algorithms both use 32 playouts per move.


\begin{table}[h]
  \centering
  \caption{Playing 400 games with different FPU and different numbers of evaluations against the $\mu$ FPU,  $\frac{\sigma}{\sqrt{n}} < 0.0250$.}
  \label{tableFPU}
  \begin{tabular}{lrrrrrrrrrrrrrrrrrr}
FPU       &     32 &     64 &    128 &    256 & 512 \\
constant  & 0.1300 & 0.1300 & 0.0475 & 0.0175 & \\
best mean & 0.3775 & 0.3450 & 0.3275 & 0.3150 & 0.2725 \\
  \end{tabular}
\end{table}

It is clear that the $\mu$ FPU is the best option. In the remainder of the experiments we use the $\mu$ FPU.

\subsection{Trees and Transposition Table}

We now experiment with using a plain tree associated to a transposition table. An entry in the transposition table only contains the evaluation and the prior. A node in the tree contains the children and the statistics of the state. We compare it to PUCD, i.e. the PUCT algorithm with a transposition table that stores both the statistics, the evaluation and the priors. PUCD searches with a Directed Acyclic Graph while its opponent develops a plain tree with transpositions only used to initialize leaves.

Table \ref{tablePUCD} gives the budget used by each algorithm (the number of forward of the neural network), the number of descents of the plain tree algorithm using this budget and the ratio of the number of descents divided by the number of forwards and the win rate of the plain tree algorithm. Both PUCD and the plain tree algorithm are called with a batch of size one. The PUCD algorithm makes exactly as many descents as forwards when the plain tree algorithm makes more descents than forwards. The ratio of the number of descents divided by the number of forwards increases with the budget. We can see that both algorithms have close performances with the plain tree algorithm getting slightly better with an increased budget.

\begin{table}[h]
  \centering
  \caption{Playing 400 games with a tree, a transposition table and a batch of size 1 against PUCD with a batch of size 1, $\frac{\sigma}{\sqrt{n}} < 0.0250$}
  \label{tablePUCD}
  \begin{tabular}{lrrrrrrrrrrrrrrrrrr}
Budget  & Descents & Ratio & Winrate \\
256     &    273.08 & 1.067 & 0.4800  \\ 
1024    &  1 172.21 & 1.145 & 0.4875 \\ 
4096    &  5 234.01 & 1.278 & 0.5275 \\ 
  \end{tabular}
\end{table}

\subsection{The Virtual Mean}

In order to compare the virtual loss and the Virtual Mean we make them play against the sequential algorithm. They both use batch MCTS. The results are given in table \ref{tablePenalties}. The first column is the penalty used, the second column is the value of $vl$ the number of visits to add for the penalty used. The third column is the number of batches and the fourth column the size of the batches.  The fifth column is the average number of nodes of the tree. The sixth column is the average of the number of useful inferences made per batch. The number of inferences made can be smaller than the batch size since the batch is not always fully filled at each call of the GetBatch algorithm. The last column is the win rate against the sequential algorithm using 64 batches of size 1. All experiments are made with the maximum number of descents $N = 500$. It is normal that the number of inferences per batch is smaller than the batch size since for example the first batch only contains one state because the priors of the root are not yet known.

The best result for the Virtual Loss is with $vl = 3$ when using 8 batches. It scores 20.75\% against sequential PUCT with 64 state evaluations. The Virtual Mean with 8 batches has better results as it scores 31.00\% with $vl = 3$ against the same opponent.

We also tested the virtual loss and the Virtual Mean for a greater number of batches. For 32 batches of size 32 (i.e. inferences on a little less than 1024 states) the best result for the virtual loss is with $vl = 2$ with an average of 157.69 nodes in the tree and a percentage of 79.00\% of wins against sequential PUCT with 64 state evaluations. The Virtual Mean with $vl = 1$ and the same number of batches is much better: it  has on average 612.02 nodes in the tree and a percentage of wins of 97.00\% of its games.

In the remaining experiments we use the Virtual Mean.

\begin{table}[h]
  \centering
  \caption{Playing 400 games with the different penalties against sequential PUCT with 64 state evaluations. $\frac{\sigma}{\sqrt{n}} < 0.0250$}
  \label{tablePenalties}
  \begin{tabular}{lrrrrrrrrrrrrrr}
Penalty       & $vl$ &  B & Batch &  Nodes & Inference & Winrate\\
  &     &   &     &   &      &  \\
Virtual Loss  &    1 &  8 &    32 &  24.47 &     23.17 &  0.1300\\
Virtual Loss  &    2 &  8 &    32 &  24.37 &     24.46 &  0.1525\\
Virtual Loss  &    3 &  8 &    32 &  24.11 &     25.16 &  \bf 0.2075\\
Virtual Loss  &    4 &  8 &    32 &  23.87 &     25.53 &  0.2025\\
Virtual Loss  &    5 &  8 &    32 &  23.91 &     25.72 &  0.1600\\
  &     &   &     &   &      &  \\
Virtual Loss  &    1 & 32 &    32 & 166.09 &     28.08 &  0.7725\\
Virtual Loss  &    2 & 32 &    32 & 157.69 &     28.25 &  \bf 0.7900\\
Virtual Loss  &    3 & 32 &    32 & 151.02 &     28.30 &  0.7800\\
Virtual Loss  &    4 & 32 &    32 & 144.45 &     28.19 &  0.7550\\
  &     &   &     &   &      &  \\
Virtual Mean  &    1 &  8 &    32 &  46.45 &     20.22 &  0.2625\\
Virtual Mean  &    2 &  8 &    32 &  43.75 &     21.64 &  0.3025\\
Virtual Mean  &    3 &  8 &    32 &  41.63 &     22.10 &  \bf 0.3100\\
Virtual Mean  &    4 &  8 &    32 &  40.41 &     22.53 &  0.2400\\
  &     &   &     &   &      &  \\
Virtual Mean  &    1 & 32 &    32 & 612.02 &     26.63 & \bf 0.9700 \\
Virtual Mean  &    2 & 32 &    32 & 619.07 &     27.83 & 0.9675 \\
Virtual Mean  &    3 & 32 &    32 & 593.91 &     28.20 & 0.9500 \\
  \end{tabular}
\end{table}

\subsection{The Last Iteration}

Table \ref{tableIteration} gives the result of using the Last Iteration heuristic with different values for U. The column $vll$ contains the value of $vl$ used for the Last Iteration. We can see that the win rates are much better when using 8 batches than for table \ref{tablePenalties} even for a small U. A large virtual loss ($vll$) of 3 makes more descents but it is less accurate. Using a virtual loss of 1 is safer and gives similar results.

When using 32 batches against 512 states evaluations the win rate increases from 62.75\% for $U=0$ to 68.00\% for $U=40$.

\begin{table}[h]
  \centering
  \caption{Playing 400 games with the Last Iteration algorithm against sequential PUCT with $P$ state evaluations. $\frac{\sigma}{\sqrt{n}} < 0.0250$.}
  \label{tableIteration}
  \begin{tabular}{lrrrrrrrrrrrrrrrrrr}
U  & vl & vll &  B & Batch &   P &    Nodes & Inference & Winrate\\
10 & 3  &   3 &  8 &    32 &  64 &  109.02 &     21.90 &  0.4975\\
10 & 3  &   1 &  8 &    32 &  64 &   75.09 &     22.04 &  0.4450\\
40 & 3  &   3 &  8 &    32 &  64 &  232.09 &     22.17 &  0.5100\\
40 & 3  &   1 &  8 &    32 &  64 &  129.90 &     22.02 &  0.5275\\
 0 & 1  &   1 & 32 &    32 & 512 &  729.41 &     28.88 &  0.6275\\
40 & 1  &   3 & 32 &    32 & 512 &  962.84 &     28.81 &  0.6650\\
40 & 1  &   1 & 32 &    32 & 512 &  835.07 &     28.86 &  0.6800\\
  \end{tabular}
\end{table}

\subsection{The Second Move Heuristic}

Table \ref{tableSecondMove} gives the winrate for different budgets when playing PUCT with the second move heuristic against vanilla PUCT. We can see that the Second Move heuristic consistently improves sequential PUCT.

\begin{table}[h]
  \centering
  \caption{Playing 400 games with the second move heuristic used at the root of sequential PUCT against sequential PUCT.  $\frac{\sigma}{\sqrt{n}} < 0.0250$.}
  \label{tableSecondMove}
  \begin{tabular}{lrrrrrrrrrrrrrrrrrr}
Budget  & Winrate \\
32      & 0.5925 \\
64      & 0.6350 \\
128     & 0.6425 \\
256     & 0.5925 \\
512     & 0.6250 \\
1024    & 0.5600 \\
  \end{tabular}
\end{table}

\subsection{Ablation Study}

The PUCT constant $c=0.2$  that we used in the previous experiments was fit to the sequential PUCT on a DAG with 512 inferences. In order to test the various improvements we propose to fit again the $c$ constant with all improvements set on. The results of games against sequential PUCT for different constants is given in table \ref{tableConstants}. The $c=0.5$ constant seems best and will be used in the ablation study.

\begin{table}[h]
  \centering
  \caption{Result of different constants for 32 batches of size 32 against sequential PUCT with 256 state evaluations and a constant of 0.2. $\frac{\sigma}{\sqrt{n}} < 0.0250$.}
  \label{tableConstants}
  \begin{tabular}{lrrrrrrrrrrrrrrrrrr}
$c$ & Winrate\\
0.2 &  0.7575\\
0.3 &  0.7925\\
0.4 &  0.8100\\
0.5 &  0.8275\\
0.6 &  0.7975\\
0.7 &  0.7700\\
0.8 &  0.7550\\
1.6 &  0.5900\\
  \end{tabular}
\end{table}

Table \ref{tableLeaveOneOut} is an ablation study. It gives the scores against sequential PUCT with 512 evaluations of the different algorithms using 32 batches with some heuristics removed.

Removing the Virtual Mean heuristic is done by replacing it with the virtual loss heuristic. However the virtual loss combined with the Last Iteration is catastrophic. So we also removed both the Virtual Mean and the Last Iteration heuristics in order to evaluate removing the Virtual Mean.

Removing the $\mu$ FPU was done replacing it by the best mean FPU. The Last Iteration uses $vll=1$ and $U = 40$.

We can observe in Table \ref{tableLeaveOneOut} that all the heuristics contribute significantly to the strength of the algorithm. The Virtual Mean has the best increase in win rate, going from 29.50\% for the virtual loss to 68.00\% when replacing the virtual loss by the Virtual Mean. The Second Move heuristic also contributes to the strength of Batch MCTS.

\begin{table}[h]
  \centering
  \caption{Playing 400 games with the different heuristics using 32 batches of size 32 against sequential PUCT with 512 state evaluations. $\frac{\sigma}{\sqrt{n}} < 0.0250$.}
  \label{tableLeaveOneOut}
  \begin{tabular}{lrrrrrrrrrrrrrrrrrr}
$\mu$ FPU & Virtual Mean & Last Iteration & Second Move & Winrate\\
        y &           y  &              y &           y & 0.6800\\
        n &           y  &              y &           y & 0.4775\\
        y &           n  &              y &           y & 0.0475\\
        y &           n  &              n &           y & 0.2950\\
        y &           y  &              n &           y & 0.6275\\
        y &           y  &              y &           n & 0.5750\\
  \end{tabular}
\end{table}

\subsection{Inference Speed}

Table \ref{tableBatches} gives the number of batches per second and the number of inferences per second for each batch size. Choosing batches of size 32 enables to make 26 times more inferences than batches of size 1 while keeping the number of useful inferences per batch high enough.

\begin{table}[h]
  \centering
  \caption{Number of batches per second according to the size of the batch with Tensorflow and a RTX 2080 Ti.}
  \label{tableBatches}
  \begin{tabular}{lrrr}
Size & Batches per second & Inferences per second\\
1    & 38.20 &    38\\
2    & 36.60 &    73\\
4    & 36.44 &   146\\
8    & 33.31 &   267\\
16   & 32.92 &   527\\
32   & 31.10 &   995\\
64   & 26.00 & 1 664\\
128  & 18.32 & 2 345\\
  \end{tabular}
\end{table}

\section{Conclusion}

We have proposed to use a tree for the statistics and a transposition table for the results of the inferences in the context of batched inferences for Monte Carlo Tree Search. We found that using the $\mu$ FPU is what works best in our framework. We also proposed the Virtual Mean instead of the Virtual Loss and found that it improves much Batch MCTS. The Last Iteration heuristic also improves the level of play when combined with the Virtual Mean. Finally the Second Move heuristic makes a good use of the remaining budget of inferences when the most simulated move cannot be replaced by other moves.


\bibliographystyle{IEEEtran}
\bibliography{main}

\end{document}